\documentclass[11pt]{article}
\usepackage{acl2013}
\usepackage{times}
\usepackage{latexsym}
\usepackage{amsmath}
\usepackage{amsfonts}
\usepackage{multirow}
\usepackage{url}
\usepackage{xspace}
\usepackage{verbatim}
\usepackage{algorithm2e}
\usepackage[retainorgcmds]{IEEEtrantools}

\usepackage{color}
\usepackage[pdftex]{graphicx}
\usepackage{caption}
\usepackage{subcaption}
\usepackage{multirow}

\numberwithin{equation}{section}

\newcommand{\ignore}[1]{}
\def\Tiny{ \font\Tiny = cmr10 at 7pt \relax  \Tiny}

\title{``Translation can't change a name'': \\Using Multilingual Data for
Named Entity Recognition}

\author{Manaal Faruqui\\
  Language Technologies Institute \\
  Carnegie Mellon University \\
  Pittsburgh, PA, 15213, USA \\
  {\tt mfaruqui@cs.cmu.edu} 
}
\date{}

\begin{document}
\maketitle

\begin{abstract}
Named Entities (NEs) are often written with no orthographic changes across different languages that share a common alphabet. We show that this can be leveraged so as to improve named entity recognition (NER) by using unsupervised word clusters from secondary languages as features in state-of-the-art discriminative NER systems. We observe significant increases in performance, finding that person and location identification is particularly improved, and that phylogenetically close languages provide more valuable features than more distant languages.
\end{abstract}

\section{Introduction}
\label{sec:intro}

Named Entity Recognition is an important pre-processing step for many NLP tasks. As is often the
case for NLP tasks, most of the work has been done for English. A major reason for this is
the unavailability of manually labeled training data for these languages. 
In this paper, we address this problem by showing that even in the absence of huge amounts
of training data for a given language, 
unlabeled data from the same language 
as well as from other languages can be used to improve the existing NER systems. Our approach follows a
semi-supervised setting, where, in addition to a small amount of training data, we assume availability
of large amounts of unlabeled data from a couple of other languages that are written in a similar or same script as  
the target language. For example, we hypothesize that data from \textit{German} \& \textit{English} can be 
used interchangeably to train NER systems for each other.

Our hypothesis stems from the fact that NEs behave similarly across languages~\cite{green2011entity}, more so,
some of the NEs like the names of the locations and people need not even undergo any orthographic
transformation while being used in different languages. For example, \textit{Barack~Obama} is spelled the same across all the languages that use roman script 
like English, German and French (cf. Table~\ref{tab:example}). We leverage this repetitive information from different languages and show that it can be used to improve the performance of NER system for a given language.

\begin{table}[tb]
  \centering
  \small
  \begin{tabular}{p{1cm}p{6cm}}
    \hline \hline
    English & The \textbf{Obama} administration has poured billions of dollars into $\dots$ \\ 
    German & \textbf{Barack Obama} hat 2012 mit dieser Strategie die Pr\"{a}sidentschaftswahlen gewonnen. \\ 
	French &  $\dots$ et en tentant de faire dérailler les nouvelles règles prudentielles, ce démocrate s'est mis à dos \textbf{Barack Obama}.\\
	\hline
	\end{tabular}
   \caption{Occurrence of \textbf{Barack Obama} across three languages.}
  \label{tab:example}
\end{table}

In addition to using manually labeled data for training, we use word clusters obtained from a large monolingual corpus using unsupervised clustering. Word clustering is widely used to reduce the 
number of parameters in statistical models which leads to improved generalization 
\cite{Brown:1992:CNG:176313.176316,kneser-ney-91,koo-08}. Word clusters can effectively capture syntactic, semantic, or distributional 
regularities among the words belonging to the group~\cite{turian:2010}.
We acquire such semantic and syntactic similarities from large, unlabeled corpora (\S\ref{sec:setup}) that can support the 
generalization of predictions to new, unseen words in the test set while avoiding overfitting.
We show improvements in the NER system performance when informed with these unsupervised clusters
for a number of languages (\S\ref{sec:mono}) as well as from noisy twitter data (\S\ref{sec:multi}).

\section{Methodology}

Our methodology of using secondary language data to train NER for the target language consists of two steps:
(1) Training word clusters from unlabeled secondary language data, (2) Use the word clusters as features in 
training the target language NER along with labeled target language data. 

Since named entities are not a closed word class,
it is highly probable that during the test time a named entity is encountered which was not present in the training data.
To encounter this sparsity problem, word clusters (trained on a large unlabeled corpus) are used as features in the sequence tagging problems~\cite{Clark:2003:CDM:1067807.1067817,faruqui10:_training,tackstrom2012cross}.
Thus, an unseen named entity might belong to the same word cluster as some of the seen entities which reinforces
the classifier's belief that it is indeed a named entity, improving the classifier's performance.

However, the intuition behind using
secondary language word clusters as features is that often proper nouns like names of people or locations
are spelled the same across orthographically similar languages. Hence, an unseen named entity in the test set
might have been present in the word clusters generated from an unlabeled corpus of a secondary language.

\section{Experimental Setup}
\label{sec:setup}

\paragraph{Tools:} We use Stanford Named Entity Recognition system\footnote{\url{http://nlp.stanford.edu/software/CRF-NER.shtml}} 
which uses a linear-chain Conditional Random Field to predict the most likely sequence of NE 
labels~\cite{Finkel:2009:NNE:1699510.1699529}. It uses a variety
of features, including the word, lemma, and POS tag of the current word and its context, n-gram features,
and word shape. This system supports inclusion of distributional similarity
features such as the ones that we want to use in the form of word clusters. 
%These features measure
%how similar a token is to another in terms of its occurrences in the document and can help in classifying previously 
%unseen words, under the assumption that strong semantic similarity corresponds to the same named entity 
%class.

For word clustering, we use the~\cite{Clark:2003:CDM:1067807.1067817} 
system\footnote{\url{http://www.cs.rhul.ac.uk/home/alexc/pos2.tar.gz}} which in addition to the standard distributional 
similarity features also uses morphological information about a word using a character-based HMM model for identifying 
similar words . This gives it the capability to more easily cluster unknown words in morphologically complex languages 
like German as compared to only the distributional similarity based approaches~\cite{Brown:1992:CNG:176313.176316}.

\paragraph{Data:} We evaluate our approach on four different languages namely: \textit{German, English, Spanish} \& \textit{Dutch}.
The training and test datasets for German and English were obtained from the shared task 
``Language Independent Named Entity Recognition''\footnote{\url{http://www.cnts.ua.ac.be/conll2003/ner/}} at CoNLL 2003~\cite{tjongkimsang2003conll}. 
%The German data is a collection of articles from the Frankfurter Rundschau newspaper and the English data is a 
%collection of news wire articles from the Reuters corpus. 
The training and test data for Dutch and Spanish were 
obtained from a similar shared task\footnote{\url{http://www.cnts.ua.ac.be/conll2002/ner/}} at CoNLL 
2002~\cite{tksintro2002conll}. 
%The Spanish data is a collection of newswire articles made available by the Spanish EFE 
%News Agency and the Dutch data consist of four editions of the Belgian newspaper ``De Morgen''. 
The training data is 
annotated with four entity types: \textit{person} (PER), \textit{location} (LOC), \textit{organisation} (ORG) and 
\textit{miscellaneous} (MISC).

For generalization data we use the news commentary corpus released by 
WMT-2012~\footnote{\url{http://www.statmt.org/wmt12/translation-task.html}} containing articles from 2011. It 
contains monolingual corpora for \textit{English, German, French, Spanish} and \textit{Czech}. Each of these corpora
on an average have approximately $200$ million tokens. Although this corpora collection doesn't include a corpus for 
Dutch, we do not search for any other source of Dutch, because our aim in the first place is to show 
cross-language utility of resources. We train clusters of size $400$ for each language as this is suitable number for the
size of our generalization corpus \cite{faruqui10:_training}.

\section{Results}

\begin{table*}[tb]
  \centering
  \small
  \begin{tabular}{l|c|c|c|c}
    Word Clusters & English NER & German NER & Dutch NER & Spanish NER \\ 
	\hline
	Baseline (None) & 91.67 & 68.27 & 79.38 & 79.94 \\
	\hline
	English & $\textbf{93.28}^{**}$ & $71.48^{**}$ & $81.65^{**}$ & $80.57^{*}$ \\ 
    German & $92.92^{**}$ & $\textbf{77.16}^{**}$ & $\textbf{81.88}^{**}$ & $80.4$ \\ 
	Spanish & $92.43^{*}$ & $70.96^{**}$ &  $81.33^{**}$ & $\textbf{81.57}^{**}$\\
    French & $92.36^{*}$ & $71.04^{**}$  & $81.63^{**}$ & $80.63^{*}$\\ 
	Czech & $92.44^{*}$ & $71.47^{**}$ & $80.71^{**}$ & $80.62^{*}$ \\ 
	%\hline
	%English + German & & & &\\
	%French + Spanish  & & & &\\
	\hline
	Average & $92.69^{**}$ & $72.42^{**}$ & $80.86^{*}$ & $81.34^{**}$ \\
   \end{tabular}
   \caption{$F_1$ scores of NER systems trained across languages. Values in bold indicate highest improvement for that NER. Statistical significance: **  $\rightarrow (p<0.01)$ and * $\rightarrow (p<0.05)$.}
  \label{tab:trNER}
\end{table*}

\subsection{Training on monolingual clusters}
\label{sec:mono}
%\paragraph{Training on monolingual clusters:}
Table~\ref{tab:trNER} shows the $F_1$ score
of the NER systems when trained using the word clusters obtained from five different languages (one at a time). 
The top row in the table shows the baseline NER system performance trained only on the labeled data without
any word clusters. The best results are
obtained when word clusters of the same language are used and in 
all the cases these results are highly statistically significant ($p<0.01$) according to a McNemar's 
test~\cite{Dietterich98approximatestatistical}.

For English, using German word clusters helps improve its performance by $1.3$ $F_1$
points which is highly statistically significant. We observe a consistent significant improvement ($p<0.05$)
using word clusters from French, Spanish and Czech. 
For a highly morphologically complex language like German it turns 
out that any amount of data from other languages can be put to good use, as it can be seen that using clusters from
any language gives a highly significant improvement in the $F_1$ score. On an average, an absolute 4 points increase
in the score is noted for German. 

For Spanish, a significant improvement improvement is obtained when using
word clusters from all the languages except for German which might be contributed to the fact they are from different 
language sub-families. As we do not have generalization data for Dutch, we used word clusters from all other languages
and observed that the highest improvement is obtained when word clusters from German are used. This is expected
as German and Dutch are typographically very similar languages and come from the Germanic language family.
It is notable that even a language like Czech which is a slavic language and is significantly different from the
romance and germanic languages gives significant improvement for all the languages.

\begin{table}[tb]
  \centering
  \small
  \begin{tabular}{l|c|c}
    NER & Baseline & Multilingual \\
	\hline
	English & $91.67$ & $\textbf{93.31}^{**}$ \\ 
    German & $68.27$ & $\textbf{77.37}^{**}$ \\ 
	Dutch & $79.38$ &  $\textbf{82.15}^{**}$ \\
    Spanish & $79.94$ & $\textbf{81.74}^{**}$ \\ 
	\end{tabular}
   \caption{$F_1$ scores of NER systems trained on multilingual word clusters. %Statistical significance: *  $\rightarrow (p<0.05)$.
   }
  \label{tab:trMultiNER}
\end{table}

\subsection{Training on multilingual clusters}
\label{sec:multi}
%\paragraph{Training on multilingual clusters:} 
In events when we have unlabeled data from the given language
and many secondary languages, we should ideally be able to use word clusters from all of them. In order to do this, we
merge the word clusters from different languages together by: (1) Keeping all the words of the given language 
intact (2) Importing only those words from the secondary language which are not present in the original language (thus improving recall). 
While importing a foreign word, it is assigned to that word cluster which has maximum number of words in common with its 
present cluster. 

Using this technique we merge all the word clusters from different languages (German, English, French, Spanish \& Czech) 
together into one multilingual word clustering.
Table~\ref{tab:trMultiNER} shows the performance of multilingual word clusters trained NERs against the baseline
NERs. In all the cases, again the NERs trained with multilingual guidance
perform significantly better than the baseline NERs and also perform better than the NERs trained with
only monolingual word clusters (cf. Table~\ref{tab:trNER}).

\begin{table}[tb]
  \centering
  \small
  \begin{tabular}{l|c|c}
    NER & Baseline & Twitter clusters \\
	\hline
	English & $91.67$ & $92.61^{**}$ \\ 
    German & $68.27$ & $71.21^{**}$ \\ 
	Dutch & $79.38$ & $80.84^{**}$ \\
    Spanish & $79.94$ & $80.39$ \\ 
	\end{tabular}
   \caption{$F_1$ scores of NER systems trained using noisy twitter data. %Statistical significance: **  $\rightarrow (p<0.01)$.
   }
  \label{tab:trTwitterNER}
\end{table}

\subsection{Training on out-of-domain clusters}

The labeled NE training data for all of the languages we have used comes
from the newswire domain. Thus the news-commentary data (cf. Sec.~\ref{sec:setup}) that we use for word clustering is 
in-domain data. Since we cannot always expect to obtain in-domain data, we use word 
clusters\footnote{\url{http://www.ark.cs.cmu.edu/TweetNLP/}} obtained from a large
collection of English tweets containing approx. $850$ million tokens clustered into $1000$ classes for 
generalization~\cite{owoputi2013improved}. Table~\ref{tab:trTwitterNER} shows the
performance of NER systems trained using the twitter word clusters. For English, German and French we again obtain
a highly significant improvement; however the improvements obtained using the out-of-domain data are lesser than that
obtained using in-domain data.

\subsection{Analysis}

\begin{table}[tb]
  \centering
  \small
  \begin{tabular}{l|c|c|c|c||c}
    NEs & English & German & Dutch & Spanish & Avg. \\
	\hline
	LOC & $\textbf{1.9}$ & $2.1$ & $\textbf{4.1}$ & $0.9$ & $\textbf{2.25}$\\ 
    MISC & $0.9$ & $1.2$ & $0.1$ & $0.4$ & $0.65$\\
	ORG & $1.7$ & $1.0$ & $3.2$ & $0.1$ & $1.50$\\ 
	PER & $1.3$ & $\textbf{4.6}$ & $3.4$ & $\textbf{1.7}$ & $\textbf{2.75}$\\
	\end{tabular}
   \caption{Category-wise improvement ($\Delta F_1$) of NER systems over the baseline model.}
  \label{tab:improvNER}
\end{table}

In order to verify our hypothesis that certain NEs, specially the names of \textit{people} and \textit{locations}
might not go orthographic changes while transferring to similar languages we look at the category-wise improvement
of the NER systems when trained using word clusters of a language (other than self) that gives the best results
against the baseline model. For example, for Spanish NER we compare the baseline model with the model 
trained using French word clusters. Table~\ref{tab:improvNER} shows the cateogry-wise improvement in the $F_1$ score
of the NER systems. 

For all the languages, the best improvement is obtained for the LOC or the PER class. 
On an average, the highest improvement is obtained for PER followed by LOC and least for MISC category. The reason for
poor improvement in MISC category is that it mostly contains linguistically inflected forms of proper nouns like 
\textit{Italian}, \textit{Irish}, \textit{Palestinians} etc. which translate into different lexical forms in different 
languages.

\begin{table}[tb]
  \centering
  \small
  \begin{tabular}{c|c|c|c}
    English & German & Dutch & Spanish \\
	(German) & (English) & (German) & (French) \\
	\hline
	Schalke & Michoac\'an & Barrichello & Ledezma \\
	Qu\'ebec & Bernardin & Isr\"ael & Emerson \\
	Lattek & Yalcin & Cofinimmo & Bustamante \\
	Hartmann & Bahraini & Netanyahu & Guayana \\
	D\"usseldorf & Nolan & Tamas & Lyon \\
	\end{tabular}
   \caption{Most frequent out-of-vocabulary words present in the secondary word clusters.}
  \label{tab:oovWords}
\end{table}

We now analyse the words present in the test set which are out-of-vocabulary (OOV) of the training set. A fraction of these 
OOV words are present in the word clusters that we obtain from a different language, most of which are names of locations
or people as we hypothesised. We list the most frequent such words from the test set in Table~\ref{tab:oovWords}. 
%The cluster information of these OOV words makes it easier for the NER to make a prediction and hence helps improve its performance.

\section{Related Work}
\label{sec:relatedWork}

Our work is primarily inspired by \newcite{faruqui10:_training} which shows that a substantial improvement 
in the German NER system performance can be obtained by using unsupervised German word clusters. NER 
systems have been trained using the same technique for other languages like English 
\cite{Finkel:2009:NNE:1699510.1699529}, Croatian \cite{glavavs91croner} and 
Slovene \cite{ljubevsic2012building}.
Other approaches to enhance NER include that of transfer of linguistic structure from one language to 
another~\cite{tackstrom2012cross,faruqui2013information} by aligning word clusters across languages. 
\newcite{green2011entity} exploits the fact that NEs retain their shape across languages and
tries to group NEs across language together. 

In a broader perspective this can be framed as a problem of resource sharing~\cite{resourceSharing} among 
different languages. 
Languages that are closely related like Hindi and Urdu benefit from sharing 
resources for NLP tasks~\cite{visweswariah-chenthamarakshan-kambhatla:2010:POSTERS}. 
Also, closely
related are the problems of multilingual language learning using unsupervised and supervised 
approaches~\cite{Diab:2003:WSD:997603,Guo:2010:COM:1858681.1858837} and cross-lingual annotation
projection applied to bootstrapping parsers~\cite{hwa:bootstrapping,citeulike:4815235}. 
Multilingual guidance has also been used for training Part-of-Speech (POS) taggers~\cite{Snyder:2008:UML:1613715.1613851,Snyder:2009:AML:1620754.1620767,das2011unsupervised}.

Our approach is different from the previous approaches in that we are directly using data from secondary 
languages for training NER systems for the given language instead of deriving any indirect 
knowledge from the secondary language data using projection or bilingual clustering techniques. 
It is simple and significantly effective.

\section{Conclusion}

We have shown that a statistically significant improvement in the performance of NER system for a given language can
be obtained when the training data is supplemented with word clusters from a secondary language(s) which is written using the
same alphabet. The amount of help provided by this secondary language depends on how similar the secondary language is to the 
given language phylogenetically and also on the domain of the data from which the word clusters are obtained. This 
performance improvement occurs because many of the NEs, specially, names of 
\textit{persons} and \textit{locations} remain the same when used in a different language and hence the word class
information of such an OOV word is helpful in predicting its NE class.

\bibliography{references.bib}
\bibliographystyle{naaclhlt2013}
\end{document}